\documentclass[conference]{IEEEtran}
\IEEEoverridecommandlockouts
\usepackage{cite}
\usepackage{amsmath,amssymb,amsfonts}
\usepackage{algorithmic}
\usepackage{graphicx}
\usepackage{textcomp}
\usepackage{xcolor}
\def\BibTeX{{\rm B\kern-.05em{\sc i\kern-.025em b}\kern-.08em
    T\kern-.1667em\lower.7ex\hbox{E}\kern-.125emX}}
\begin{document}

\title{Discovering Sparse Recovery Algorithms\\ Using Neural Architecture Search
\thanks{The authors would like to acknowledge the support of the DARPA DIAL program under project HR00112490486. PY, SG, and CH would also like to gratefully acknowledge the support of New York University's HPC resources and personnel. PY was supported in part by the US Department of Education's GAANN Fellowship.}
}

\author{\IEEEauthorblockN{Patrick Yubeaton, Sarthak Gupta}
\IEEEauthorblockA{\textit{Electrical and Computer Engineering} \\
\textit{New York University}\\
New York, USA \\
wpy2004@nyu.edu}
\and
\IEEEauthorblockN{M. Salman Asif}
\IEEEauthorblockA{\textit{Electrical and Computer Engineering} \\
\textit{University of California, Riverside}\\
Riverside, USA \\
sasif@ucr.edu}
\and
\IEEEauthorblockN{Chinmay Hegde}
\IEEEauthorblockA{\textit{Tandon School of Engineering} \\
\textit{New York University}\\
New York, USA \\
chinmay.h@nyu.edu}
}

\maketitle

\begin{abstract}
The design of novel algorithms for solving inverse problems in signal processing is an incredibly difficult, heuristic-driven, and time-consuming task. In this short paper, we the idea of automated algorithm discovery in the signal processing context through meta-learning tools such as Neural Architecture Search (NAS). Specifically, we examine the Iterative Shrinkage Thresholding Algorithm (ISTA) and its accelerated Fast ISTA (FISTA) variant as candidates for algorithm ``rediscovery". We develop a meta-learning framework which is capable of rediscovering (several key elements of) the two aforementioned algorithms when given a search space of over 50,000 variables. We then show how our framework can apply to various data distributions and algorithms besides ISTA/FISTA.
\end{abstract}

\begin{IEEEkeywords}
sparse recovery, shrinkage, neural architecture search, meta-learning.
\end{IEEEkeywords}

\section{Introduction}

The focus of this paper is on sparsity-constrained inverse problems. These problems have a rich history and impact signal and image processing applications ranging including compressive sensing, super-resolution, deblurring, phase retrieval, and lensless imaging. A large ecosystem of algorithms has emerged for solving such problems over the last 20 years: interior point methods based on linear programming (LP)~\cite{donoho2006compressed}, greedy algorithms such as Frank-Wolfe, the iterative shrinkage and thresholding algorithm (ISTA~\cite{combettes2005signal}) and its fast variant (FISTA)~\cite{beck2009fast}, and the alternating directions method of multipliers (ADMM). {The development of each such method has been the result of mathematical ingenuity}, and progress in this space has been staccato and reliant on human intuition. We ask the question: 
\begin{quote}
\emph{Can a meta-algorithm automatically (re)discover sparse optimization approaches?}  
\end{quote}

To address this question, we present a framework for {automated (re)discovery of accelerated, iterative, sparsity-constrained optimization methods}. Our framework takes important steps towards rediscovering known and performant optimization algorithms such as ISTA (and FISTA). 

At a high level our framework rests on two key ideas. First, we view any iterative sparse recovery algorithm as a nonlinear, potentially time-varying dynamical system, which can be expressed by a generic ``unrolled neural network" with shared weights, residual connections, and pointwise thresholding. This viewpoint has been validated numerous times in the recent learning-to-optimize (L2O) literature~\cite{gregor2010learning,liu2019alista,chen2022learning}. 

Second, we view the discovery of efficient algorithms as a neural architecture search (NAS) problem~\cite{zoph2016neural,zoph2018learning,liu2018darts}, which is our key novel contribution. The high-level idea is that the search space is all possible subgraphs of a generic unrolled network, and the search strategy is meta-gradient descent, an evolutionary algorithm, or reinforcement learning (RL) over this search space. Efficient architecture for the forward and adjoint operations inside the unrolled network can also be learned in our framework.

The rest of the paper is organized as follows. We start with a quick primer of sparse recovery from linear observations, and some associated algorithms. We discuss the learning-to-optimize framework that solves these problems using an unrolled neural network approach, as well as give a very brief introduction to neural architecture search. We then present our approach, and showcase its performance via numerical experiments. We finally conclude with a discussion.

\begin{figure*}[t]
\centering
    \begin{tabular}{cc}        \includegraphics[width=0.3\linewidth]{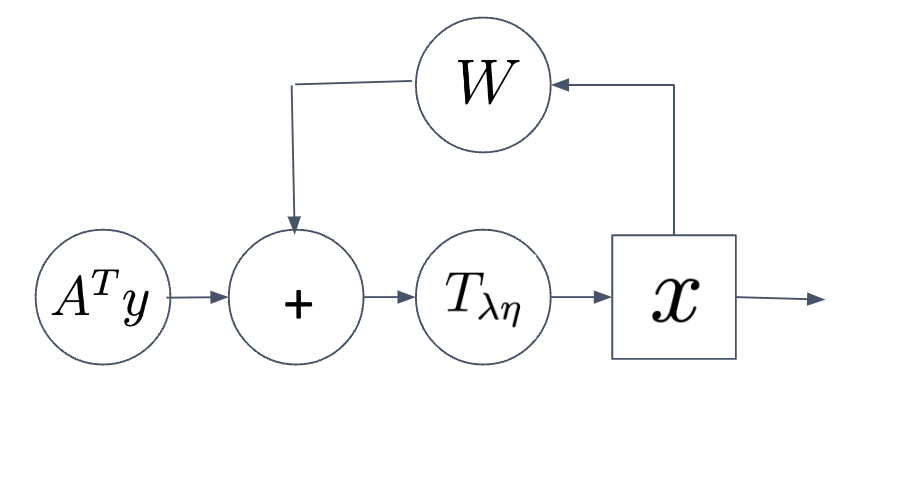}
    &
    \includegraphics[width=0.6\linewidth]{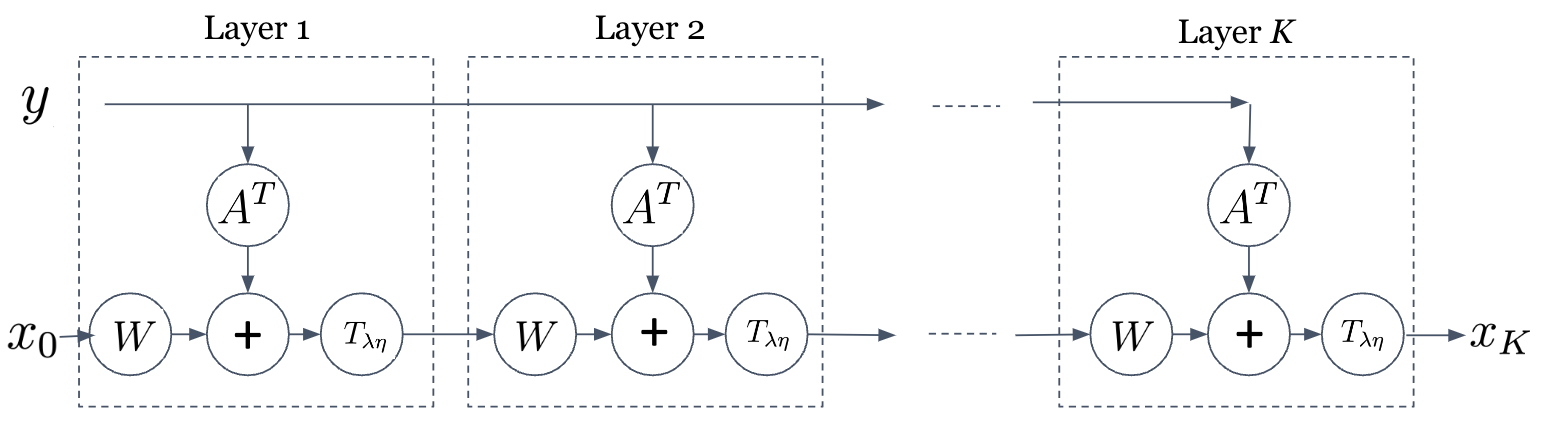} \\
    \end{tabular}        
    \caption{\sl (left) The ISTA algorithm represented via a nonlinear dynamical system, or an RNN; here, $W = I - \eta A^T A$. (right) Unfolding the RNN to $K$ stages. Our DISCO approach proposes to automatically learn the edges and weights of this recurrent neural network using NAS techniques.}
    \label{fig:ista}
\end{figure*}

\section{Background}
\subsection{ISTA and FISTA}

We specifically focus on linear inverse problems:

\begin{equation}
\min \|y - Ax\|_2^2 + \lambda\|x\|_1.
\end{equation}

These problems are central to many applications in signal processing and imaging, where the goal is to recover a sparse signal from limited or noisy measurements. There is a large ``zoo'' of algorithms for solving such problems. Popular algorithms include the Iterative Shrinkage Thresholding Algorithm (ISTA) and its accelerated variant, Fast ISTA (FISTA)~\cite{beck2009fast}. ISTA is a fundamental algorithm for solving sparse recovery problems. It operates by alternating between a gradient descent step and a soft-thresholding operation. ISTA can be represented as an RNN where each cell corresponds to one iteration of the algorithm. The basic structure of an ISTA iteration can be expressed as:

\begin{equation}
x_{k+1} = S_\lambda(x_k - \eta A^T(Ax_k - y))
\end{equation}

Here, $S_\lambda$ is the soft-thresholding operator, $\eta$ is the step size, and $A^T$ is the adjoint of the measurement matrix $A$. 

FISTA is an accelerated version of ISTA which uses principles from Nesterov's method~\cite{nesterov1983method} which introduces a momentum term to speed up convergence. 

For general linear inverse problems (without further assumptions on the linear operator $A$, such as restricted isometry~\cite{candes2008restricted}), the FISTA method achieves quadratic convergence rate, and that this rate is essentially the fastest possible (under worst case inputs).

\subsection{Learning to Optimize (L2O).} Learning-to-optimize (L2O) solvers have emerged as a framework for designing efficient iterative algorithms by learning from data, among L2O, LISTA-type solvers focusing on sparse recovery problems. LISTA~\cite{gregor2010learning} pioneered this approach by unrolling the iterations of sparse coding algorithms and optimizing their parameters through supervised learning, achieving significant acceleration compared to traditional sparse solvers. Building on this foundation, LISTA-CP~\cite{chen2018theoretical} introduced coupling constraints to enforce theoretical convergence guarantees, bridging the gap between performance and interpretability. Furthermore, TiLISTA and ALISTA~\cite{liu2019alista} improved upon LISTA by leveraging learned thresholds or adaptively adjusting step sizes, demonstrating superior generalization to unseen data. Empirical studies~\cite{chen2018theoretical} reveal that support selection strategies—where significant components are selected based on learned criteria—outperform traditional shrinkage-based methods in terms of accuracy and efficiency.

\subsection{Neural Architecture Search}

Choosing a suitable neural network architecture for complex prediction tasks such as image classification and language modeling often requires a substantial effort of trial-and-error. The sub-field of neural architecture search (NAS) addresses the problem of designing competitive architectures with as small computational budget as possible. 
Numerous approaches for neural architecture search already exist in the literature, each with their own pros and cons: these include {black-box optimization} based on reinforcement learning (RL)~\cite{zoph2016neural}, evolutionary search~\cite{real2018regularized}, and Bayesian optimization~\cite{cao2019learnable,kandasamy2018neural}. In our work below, we focus on differentiable architecture search (DARTS)~\cite{liu2018darts,cho2019one} which poses the architecture search problem in terms of subgraph discovery, and solves for an optimal subgraph using a relaxation of an integer program.

\begin{figure*}[!t]
\centering
    \includegraphics[width=\linewidth]{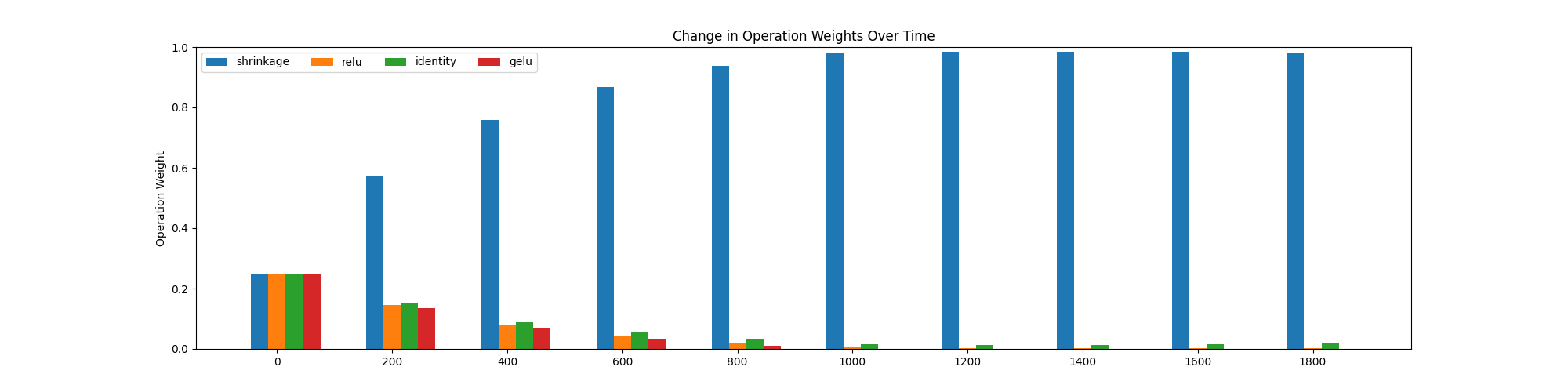}
    \caption{\sl We show the NAS weights for each operation during the model training. The NAS weights are obtained by taking the softmax of the NAS $\alpha$ values for each layer and adding up the values for each operation. We then normalize the values to be between 0 and 1.} 
    \label{fig:nas_evo_old}
\end{figure*}

\begin{figure*}[!t]
    \centering
    \includegraphics[width=0.5\linewidth]{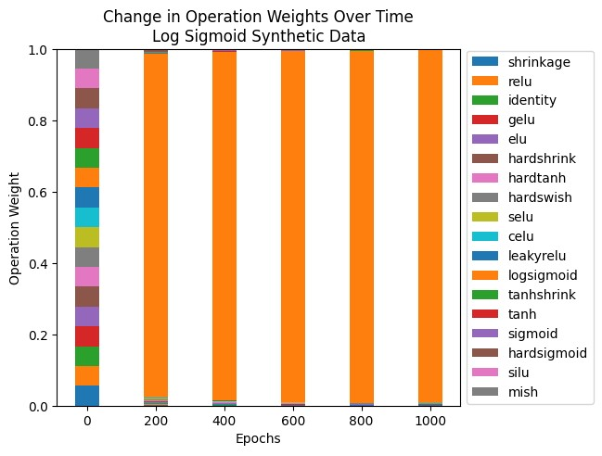}
    \caption{We show the NAS weights for each operation during model training. The NAS weights are obtained by taking the softmax of the NAS $\alpha$ values for the one layer model and normalizing the values to be between 0 and 1. Figure best viewed in color.}
    \label{fig:nas_log_sig}
\end{figure*}

\section{Our Results}

\subsection{Recovering the soft thresholding operator}

Here, we provide some initial results using our NAS framework for (re)discovering sparse recovery algorithms. 

\paragraph{Dataset}
We create a synthetic sparse coding dataset to train our model. The dataset is created with the following steps: Initialize a dictionary matrix W$\in\mathbb{R}^{50\times 200}$ where each parameter is sampled from the uniform distribution $[0.0, 1.0)$. Normalize the dictionary matrix and scale it by 10. Define a sparsity level $s=4$. Generate a set of sparse z$\in\mathbb{R}^{200 \times 1}$ vectors. Each vector has $s$ non-zero parameters which are drawn from a uniform random distribution $(-1.0, 1.0)$. Generate the set of input signals x by multiplying the z vectors to the dictionary matrix W.

\paragraph{Model Architecture}
In particular, we define an architecture where each layer's activation function is a weighted sum of multiple activation functions (ReLU, GELU, SILU, etc.). NAS takes this architecture and uses gradient descent to minimize the error that architecture produces for a given dataset. The final architecture produced by NAS only has one activation function per layer.

We want our model to learn that the shrinkage operator is the best activation function to use after the gradient step. This informed the majority of our model architecture choices as shown below. 

The input for our model is a signal x$\in\mathbb{R}^{50\times 1}$. The output for our model is the sparse vector used to reconstruct the signal: z$\in\mathbb{R}^{200\times 1}$. We initialize $z_0\in\mathbb{R}^{200\times 1}$ with all zeros. The model is given the dictionary matrix W during model creation. The model is initialized with 6000 layers. Each layer is composed of one gradient operation and a weighted sum of the activation functions. The model is given four activation functions to choose from: shrinkage, ReLU, GeLU, and identity. We calculate the squared Frobenius norm of the dictionary matrix and assign that to the parameter "c". The threshold value for the shrinkage function is $\lambda=0.01/c$. The gradient update parameter $\eta = 1/c$. 

We initialize $\alpha$ parameters for NAS to select between the four activation functions. We have one parameter per activation function per layer. Thus, $\alpha\in\mathbb{R}^{6000\times 4}$. The parameters are all initialized to 1. The model's forward pass works as follows: We perform the gradient operation given the input signal x, and our current estimate of z. We take the softmax of the current layer's $\alpha$ parameters. We perform a weighted sum of the activation functions being applied to the current estimate of z. The coefficients are the values obtained from the softmax of the $\alpha$ parameters. We repeat this process for every layer in the model.

\paragraph{Training Hyperparameters}
We generate a synthetic sparse coding dataset with 12,500 samples above. 80\% of the dataset is used for training while 20\% is used for validation. We initialize a model with 6000 layers and a learning rate of 0.05 for the $\alpha$ parameters. We optimize the $\alpha$ parameters with respect to the mean squared error using the ADAM optimizer. The model is trained for 2000 epochs before binarizing the activation functions.

The NAS has four choices for its activation function. The NAS will increase the value of $\alpha$ parameters for activation functions that decrease the loss. Therefore, we are looking for $\alpha$ values with large positive magnitudes to see which activation functions are most valued by the NAS.

\begin{figure*}[!t]
    \centering
    \includegraphics[width=\linewidth]{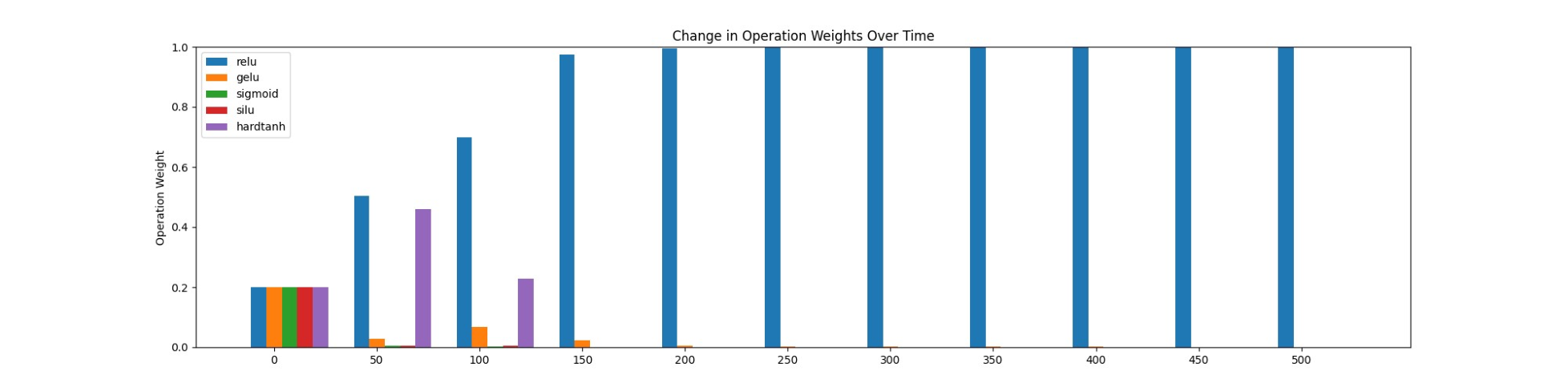}
     \includegraphics[width=\linewidth]{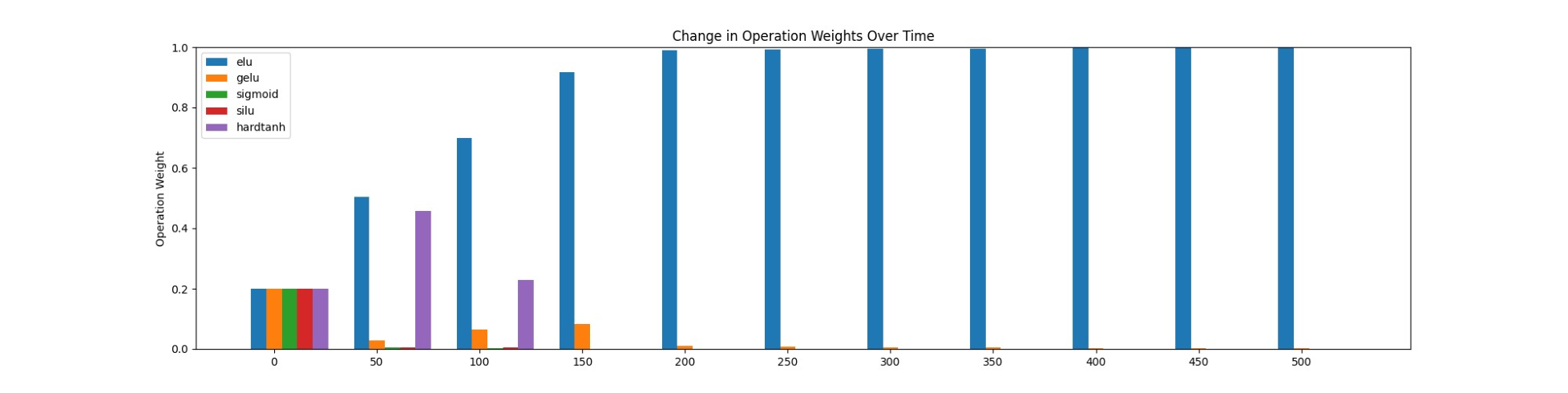}
    \caption{We show the NAS weights for each operation during model training. The NAS weights are obtained by taking the softmax of the NAS $\alpha$ values for the one layer model and normalizing the values to be between 0 and 1.}
    \label{fig:relu_nas}
\end{figure*}

\subsection{Decreasing the Training Time of NAS}
In the previous experiment, we explored the impact of increasing the NAS search space for the projection function. Our framework was still able to successfully learn the shrinkage operation, but this came with a massively increased training time. We saw that for 1000 epochs the original model (4 options) was able to confidently learn the shrinkage operation ($>$0.9 NAS weighting). However, for 1000 epochs with the larger search space (8 options) we saw that the model had only learned a 0.5 NAS weighting for shrinkage. This shows that the training time is massively increased by a larger search space, and thus motivates the pursuit of methods to decrease this training time.

One such method we explored was a "looped NAS model". In this model, we initialize a NAS cell with the projection function NAS parameters for one layer. We then reuse this NAS cell for all layers in the model instead of initializing one cell per layer. This change to the model simplifies the NAS problem by telling the framework that our algorithm is strictly iterative. However, we believe this is a reasonable assumption to make for most human-explainable algorithms. In addition, it massively lowers the parameter count allowing for quicker training time.

\subsection{Can NAS learn proximal projections tailored to specific data distributions?}
In the above experiments we have focused on sparse recovery data; this naturally leads to learning recurrent cells with shrinkage activation functions since shrinkage is the proximal operator of the L1-norm. Can this be extended to other proximal operators which stem from different data distributions? We explore this idea by generating synthetic datasets to mimic data from other proximal operators.

\paragraph{Dataset}
We initialize a synthetic sparse coding dataset as described in the previous section. Afterwards, we  discard the sparse signals and work with the data signals and dictionary matrix. Instead of using the shrinkage operation, we replace it with a different proximal operator candidate such as Tanh. We follow this iterative process for 10,000 epochs and save the final signal as our output signal. 
We follow the above methodology to generate datasets with "planted" proximal operators such as tanh, log sigmoid, ReLU, tanh-shrink, and sigmoid. Our experiments show that NAS is able to new proximal operators. As seen in Figure~\ref{fig:nas_log_sig}, for data generated according to the ``log sigmoid" data generating model, our NAS algorithm starts off with a uniform prior over the (at epoch 0), but quickly and learns to consistently recover the log-sigmoid activation function across all layers (epoch 200 and beyond). 

Is it possible for NAS to do the same thing for more naturally constructed data instead? We explore this idea by generating more natural data distributions such as positive and negative sparse data. In particular, we match the generation settings from previous sections but restrict the sparse signals to be only positive (or negative).

We know that operators such as the ReLU and the ELU promote positive values and have sparsity inducing properties. Therefore, we train a NAS model without the shrinkage operator to see if it can learn these other operators. These results are shown in Figure~\ref{fig:relu_nas}. We see that in both cases (ReLU as well as ELU), the NAS framework is able to learn the proximal operator that promotes positive sparse signals (ReLU/ELU). This provides further evidence that NAS is able to learn the ``right" proximal operator for the given data distribution.

\section{Discussion}

We conclude by discussing potential extensions of our approach to related problems. 

Our experiments have primarily focused on the iterative soft thresholding algorithm (ISTA). The accelerated version (FISTA) requires additional discovery of a momentum term. We have achieved partial results in this direction, but insufficient to report formally as part of this publication. Specifically, (re)discovery of the momentum term in FISTA requires recovering both a history term that depends on the immediate previous iterate (and no further), and a rather carefully chosen stepsize term. We can show using our NAS approach that we can reliably recover the history term (thus producing acceleration), but learning the right choice of step size is tricky. Presumably, techniques from Bayesian hyper-parameter selection~\cite{cho2019reducing} may be necessary.

Other directions of extension include sparse recovery from nonlinear observations (such as phaseless measurements~\cite{candes2015phase,hyder2019alternating}). This encompasses more challenging algorithmic settings, and may require expanding the NAS search space even further.

\bibliographystyle{unsrt}
\bibliography{refs,other-refs}
\end{document}